%% file: paper.tex
\documentclass[letterpaper, 10 pt, conference]{ieeeconf}

\IEEEoverridecommandlockouts
\usepackage{amsmath} 
\usepackage{amssymb}  
\usepackage[protrusion=false]{microtype}
\usepackage{tensor}
\usepackage{mathtools}
\usepackage{makecell}
\usepackage[colorinlistoftodos,prependcaption]{todonotes}
\usepackage{siunitx}
\usepackage{comment}
\usepackage{xcolor} 
\usepackage[normalem]{ulem}
\usepackage{esvect}
\usepackage{float}
\usepackage{lipsum}
\usepackage{multicol}
\usepackage{multirow}
\usepackage{flushend}

\usepackage{booktabs}

\title{\LARGE \bf
	M-LIO: Multi-lidar, multi-IMU odometry with sensor dropout tolerance}

\author{Sandipan Das$^{1,2}$, Navid Mahabadi$^{3}$, Maurice Fallon$^4$, Saikat
	Chatterjee$^{1}$
	\thanks{$^1$ KTH EECS, Sweden. \texttt{\{sandipan,sach\}@kth.se}\newline%
		$^2$ Scania, Sweden. \texttt{\{sandipan.das\}@scania.com}\newline%
		$^3$ Stockholm, Sweden. \texttt{n.mahabadi@gmail.com}\newline%
		$^4$ Oxford Robotics Institute, UK. \texttt{mfallon@robots.ox.ac.uk}}}

\usepackage[linesnumbered,ruled,vlined]{algorithm2e}

\usepackage{multirow}
\usepackage[normalem]{ulem}
\usepackage{xargs} 

\input{shortcuts} 

\makeatletter
\let\NAT@parse\undefined
\makeatother
\usepackage{hyperref}

\begin{document}
	
	\maketitle
	\thispagestyle{empty}
	\pagestyle{empty}
	
	We present a robust system for state estimation that fuses measurements from multiple lidars and inertial sensors with GNSS data. To initiate the method, we use the prior GNSS pose information.
	We then perform incremental motion in real-time, which produces robust motion estimates in a global frame by fusing lidar and IMU signals with GNSS translation components using a factor graph framework.
	We also propose methods to account for signal loss with a novel synchronization and fusion mechanism.
	To validate our approach extensive tests were carried out on data collected using Scania test vehicles (5 sequences for a total of $\approx$ 7 Km). From our evaluations, we show an average improvement of 61\% in relative translation and 42\% rotational error compared to a state-of-the-art estimator fusing a single lidar/inertial sensor pair.  
	
	
	
	


	\section{Introduction}
	\label{sec:introduction}

	State estimation, which is a sub-problem of Simultaneous Localization and 
	Mapping (SLAM), is a fundamental building block of autonomous navigation. To
	develop 
	robust SLAM systems, proprioceptive (IMU -- inertial measurement unit, wheel
	encoders) 
	and exteroceptive (camera, lidar, GNSS -- Global Navigation Satellite System)
	sensing 
	are fused. Existing systems have achieved robust and accurate 
	results \cite{lic-fusion,shan2021lvi,wisth2020vilens}; however, state estimation in dynamic conditions and when there is measurement loss or noisy is still challenging.
	
\begin{figure}
		\centering
		\includegraphics[width=1\columnwidth]{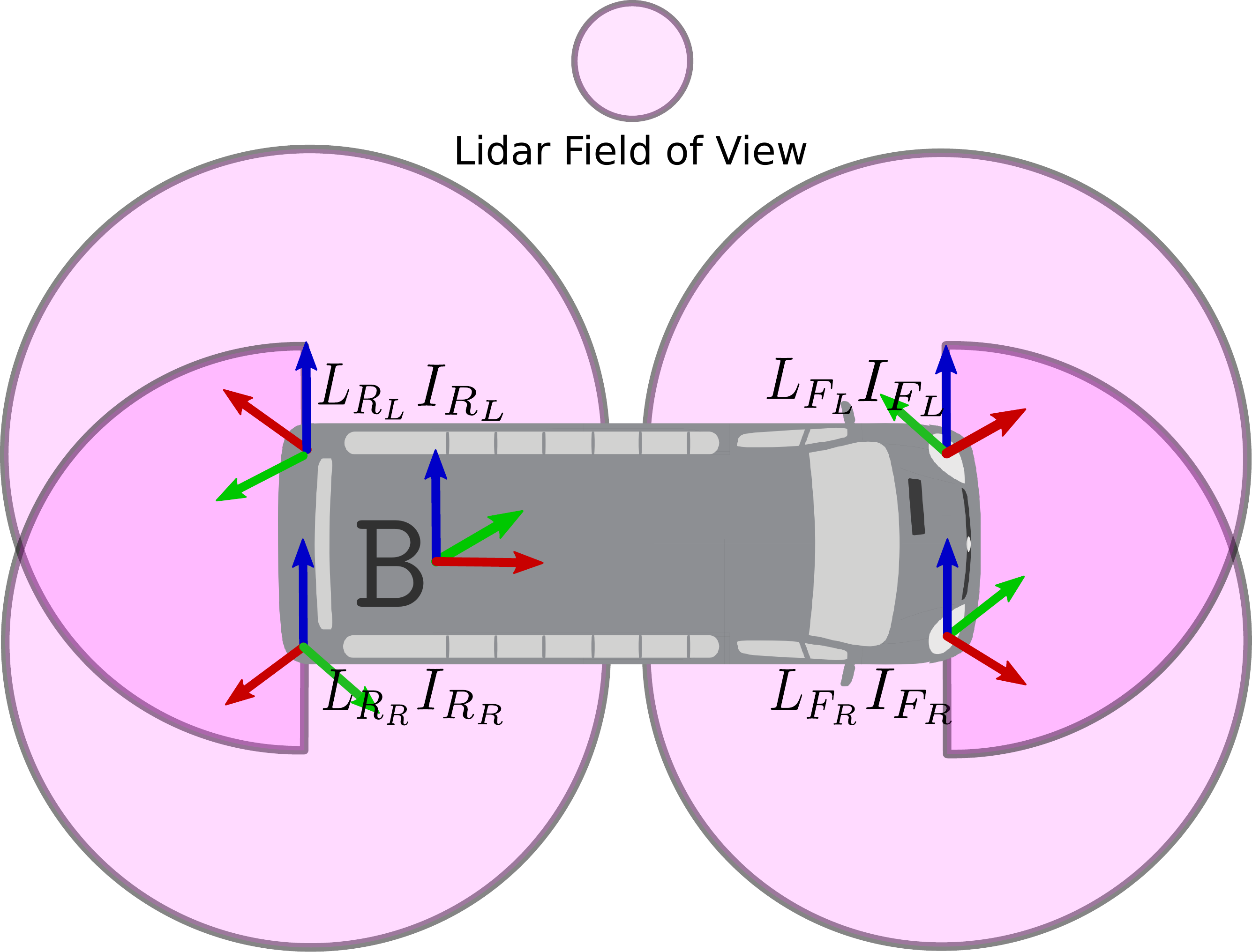}
		\caption{Illustration of the four lidars with their embedded IMUs positioned
			around the data collection vehicle. The vehicle base frame $\Base$, is located
			at the center of the rear axle. The sensor frames of the lidars are:
			$\Lidar_{{F_{R}}}$, $\Lidar_{{F_{L}}}$, $\Lidar_{{R_{R}}}$ and
			$\Lidar_{{R_{L}}}$, whereas, the sensor frames of the IMUs are:
			$\imu_{{F_{R}}}$, $\imu_{{F_{L}}}$, $\imu_{{R_{R}}}$ and $\imu_{{R_{L}}}$.
			${F_{R}}$: front-right, ${F_{L}}$: front-left, ${R_{R}}$: rear-right and
			${R_{L}}$: rear-left.} 
		\label{fig:scania-bevda}
		\vspace{-5mm}
	\end{figure}

	Compared to visual SLAM, lidar-based SLAM has higher accuracy as lidar range
	measurements 
	of up to 100m directly enable precise motion tracking. Moreover, owing to the 
	falling cost of lidars, mobile platforms are increasingly equipped with
	multiple 
	lidars \cite{sun2019scalability, geyer2020a2d2}, to give
	complementary and complete $360^{\circ}$ sensor coverage (see \Figure\ref{fig:scania-bevda}, for our data collection platform). This also improves the density of measurements which may be helpful for state estimation in degenerate scenarios such as tunnels or straight highways. We can also estimate the reliability of a state estimator by computing the covariance of measurement errors produced from multiple lidar measurements.
	
	Meanwhile, IMUs are low cost sensors which can be used to estimate a motion prior for lidar 
	odometry by integrating the rotation rate and accelerometry measurements. However, IMUs suffer from
	bias instability and are susceptible to noise. An array of multiple IMUs (MIMUs) could provide enhanced signal accuracy with bias and noise compensation as well as increasing operational robustness to sensor dropouts. Multi-lidar odometry \cite{palieri2020locus,jiao2021robust} and
	MIMU odometry \cite{skog2016inertial,faizullin2021best} have
	been studied separately; however to the best of our knowledge the fusion of the combined set has not yet been explored. 
	
	\subsection{Motivation}
	
	%
	%
	
	As our vehicles are equipped with a GNSS system, we perform the state estimation in global frame and take advantage of it to limit the drift rate. 
	Since GNSS information is unreliable in urban environments (`urban canyon') or in underground scenarios (such as mining) we also need to fuse information from onboard sensors (which might be susceptible to signal loss) to create robust state estimates. 

	To achieve this, we have identified two broad problems:
	\textit{Problem 1:} State estimation over 
	long time horizons with onboard sensing inherently suffers from drift. 
	\textit{Problem 2:} Sensor signals are susceptible to loss, due to
	networking or operational issues, which might affect reliability of the state estimates using onboard sensing.

	In our work we have addressed \textit{Problem 1}) 
	by fusing the onboard state estimator with GNSS-based estimates. For \textit{Problem 2}) we provide our formulation for the fusion of 
	multiple lidars and MIMU to create robust signals under noisy 
	conditions, which can give robustness both to failure of an individual lidar or IMU sensor as well as expanding the lidar field-of-view.
	

	
	\subsection{Contribution}
	\label{sec:contribution}
	
	Our work is motivated by the broad literature in lidar and IMU based state
	estimation. Our proposed contributions are:
	\begin{itemize}
		\item Multi-lidar odometry using a single fused local submap and handling lidar dropout scenarios.
		\item MIMU fusion which compensates for the Coriolis effect and accounts for
		potential signal loss.
		\item A factor graph framework to jointly fuse multiple lidars, MIMUs and GNSS
		signals for robust state estimation in a global frame.
		\item Experimental results and verification using data collected from Scania vehicles with the sensor setup shown in Fig. \ref{fig:coordinate-frames},
		with FoV schematics similar to Fig. \ref{fig:scania-bevda}.
	\end{itemize}
	
	\section{Related Work}
	\label{sec:related-work}
	
	There have been multiple studies of
	lidar and IMU-based SLAM after the seminal LOAM paper by
	\etalcite{Zhang}{Zhang2017} which is itself motivated by generalized-ICP work by
	\etalcite{Segal}{segal2009generalized}. In our discussion we briefly review the relevant literature.

	\subsection{Direct tightly coupled multi-lidar odometry}
	To develop real-time SLAM systems simple edge and plane features are often extracted
	from the point clouds and tracked between frames for computational efficiency
	\cite{wisth2020vilens, shan-legoloam, tixiao2020lio-sam}. Using IMU
	propagation, motion priors can then be used to enable matching of point cloud features between
	key-frames. However, this principle cannot be applied to featureless
	environments. Hence, instead of feature engineering, the whole point cloud is often
	processed which has an analogy to processing the whole image in visual odometry methods such as LSD-SLAM,
	\cite{engel2014lsd}, which is known as direct estimation. To support
	direct methods, recently \etalcite{Xu}{xu2022fast} proposed the \textit{ikd-tree} in
	their Fast-LIO2 work which efficiently inserts, updates, searches and filters 
	to maintain a local submap. The \textit{ikd-tree} achieves its efficiency
	through ``lazy delete'' and ``parallel tree re-balancing'' mechanisms. 

	Furthermore, instead of point-wise operations the authors of Faster-LIO
	\cite{bai2022faster} proposed voxel-wise operations for point cloud association
	across frames and reported improved efficiency. In our work we also maintain
	an \textit{ikd-tree} of the fused lidar measurements and tightly couple the
	relative lidar poses, IMU preintegration and GNSS prior in our propose  estimator. 
	Since, we jointly estimate the state based on the residual cost function built
	upon the multiple modalities we consider this a tightly coupled system.

	While there have 	been many studies of state estimation using single lidar and IMU, there is limited
	literature available for fusing multi-lidar and MIMU systems. Our idea
	closely resembles M-LOAM \cite{jiao2021robust}, where state estimation with
	multiple lidars and calibration was performed. However, the working principles are different as
	M-LOAM is not a direct method and MIMU system is not considered in their work. 

	Finally, most authors do not address how to achieve reliability in situations of signal loss
	--- an issue which is important for practical operational scenarios.
	
	\section{Problem Statement}
	\label{sec:problem-statement}

	\subsection{Sensor platform and reference frames}
	\label{sec:sensor-platform}
	\begin{figure}
		\centering
		\vspace{2mm}
		\includegraphics[width=1\columnwidth]{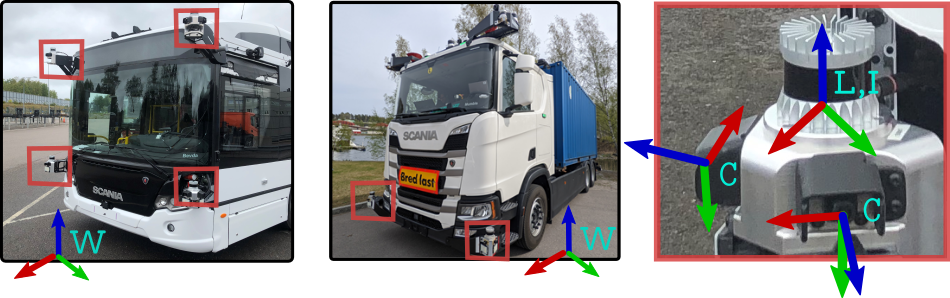} 
		\vspace{-8mm}
		\caption{Reference frames conventions for our vehicle platform. The world frame
			$\World$ is a fixed frame, while the base frame $\Base$, as shown in \Figure
			\ref{fig:scania-bevda}, is located at the rear axle center of the vehicle.
			Each
			sensor unit contains the two optical frames $\Camera$, an IMU frame, $\Imu$,
			and
			lidar frame $\Lidar$. The cameras are shown for illustration only and they are
			not used for this work.} 
		\label{fig:coordinate-frames}
		\vspace{-5mm}
	\end{figure}
	The sensor platform with its corresponding reference frames is shown in \Figure
	\ref{fig:coordinate-frames} along with the
	illustrative sensor fields-of-view in \Figure\ref{fig:scania-bevda}. Each of the sensor housings contain a lidar with a corresponding embedded IMU and two cameras. Although we do not use the cameras in this work they are illustrated here to show the full sensor setup. We used logs from a bus and a truck with similar sensor housings for our experiments. The two lower mounted modules from the rear, present in both the vehicles, are not shown here in the picture. The embedded IMUs within the lidar sensors are used to form the MIMU setup. 
	
	Now we describe the necessary notation and reference frames used in our system according to the convention of Furgale~\cite{furgale2014representing}. The vehicle base frame, $\Base$ is located on the center of the rear-axle of the vehicle. Sensor readings from GNSS, lidars, cameras and IMUs are represented in their respective sensor frames as $\GNSS$, $\Lidar^{(k)}$, $\Camera^{(k)}$ and $\Imu^{(k)}$ respectively. Here, $k \in [F_L, F_R, R_L, R_R]$ denotes the location of the sensor in the vehicle corresponding to front-left, front-right, rear-left and rear-right respectively. The GNSS measurements are reported in world fixed frame, $\World$ and transformed to $\Base$ frame by performing a calibration routine outside the scope of this work.
	In our discussions the transformation matrix is denoted as, $\T =
	\left[\begin{array}{ll}\mathbf{R}_{3\times3} & \mathbf{t}_{3\times1} \\
		\mathbf{0}^{\top} & 1\end{array}\right]\in \SEthree$ and
	$\R\R^T=\mathbf{I}_{3\times3}$, since the rotation matrix is orthogonal. 
	
	\subsection{Problem formulation}
	Our primary goal is to estimate the position
	$\tensor[_\world]{\tran}{_{\world\base}}$, orientation
	$\tensor[_\world]{\R}{_{\world\base}}$, linear velocity
	$\tensor[_\World]{\vel}{_{\world\base}}$, and angular velocity
	$\tensor[_\World]{\rotvel}{_{\world\base}}$, of the base frame $\Base$, relative
	to the fixed world frame $\World$. Additionally, we also estimate the MIMU
	biases $\tensor[_\Base]{\bias}{^{g}},\;\tensor[_\Base]{\bias}{^{a}}$ expressed
	in $\Base$ frame, as that is where it can be sensed. Hence, our estimate of vehicle's state $\State_i$ at
	time $t_i$, is denoted as:
	\begin{align}
		&\State_i =
		\left[{\R}_i,\;{\tran}_i,\;{\vel}_i,\;{\rotvel}_i,\;\bias^{a}_i,\;\bias^{g}_i\right]
		\in \SEthree \times \Real^{15},
	\end{align}
	where, the corresponding measurements are in the frames mentioned above. 

	
	\section{Methodology}
	\label{sec:methodology}
	
	\subsection{Initialization}	
    To provide an initial pose we use the GNSS measurements, $\tensor[_\World]{\T}{_{\mathtt{{\World\GNSS}}}}$ and
	determine an initial estimate of the starting yaw and translation component,
	$\tensor[_\World]{\tran}{_{\mathtt{{\World\Base}}}}$, 
	in UTM co-ordinates. The GNSS
	unit also provides IMU measurements in $\GNSS$ frame using its internally
	embedded IMU, which is transformed to $\Base$ frame. For gravity alignment we use the equations 
	\cite[Eq. 25, 26]{pedley2013tilt} to estimate roll and pitch after collecting IMU data when the vehicle is static for $t$ seconds according to the GNSS estimate.
	Subsequently, only 	the translation component $\tensor[_\World]{\tran}{_{\World\GNSS}}$, from the GNSS measurements are used. 

	Meanwhile, the noise processes and starting bias estimates for the embedded IMU sensors 
	were characterized in advance by estimating the Allan Variance 
	\cite{allan1966statistics} parameters using logs collected while stationary.

	\subsection{Handling lossy signal scenario}
\begin{figure}[!h]
		\centering
		\includegraphics[width=1\columnwidth]{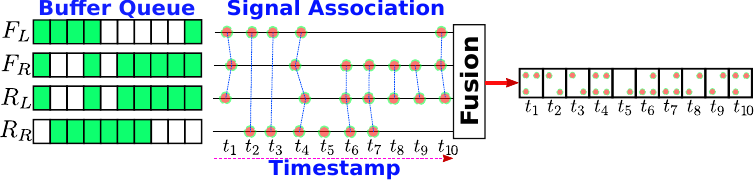}
		\vspace{-5mm}
		\caption{Handling and fusing lossy signals. Green cells denote available
			data while white cells denote a lost signal for the corresponding timestamp in
			the buffer queue.} 
		\label{fig:lossy-signal-handling}
		\vspace{-5mm}
	\end{figure}

	To synchronizate the different signals and to handling signal drop we maintain individual buffer
	queues for each of the lidar and IMU signals. We read the data from the buffers
	in a FIFO approach and compare the header timestamps (within a threshold of 10
	\si{ms} for lidar and 1 \si{ms} for IMU) to associate corresponding signals
	across different sensors as illustrated in \Figure
	\ref{fig:lossy-signal-handling}. We add an age penalty for
	unused messages in the buffer queues to avoid unbounded queue growth. Unlike the default approximate time sync policy (in ROS) we
	ensure that data from a single sensor is returned if the other corresponding
	signals are available. As an illustration, in \Figure
	\ref{fig:lossy-signal-handling}, let us assume we receive lossy lidar and IMU
	signals from our sensor setup. The green cells represent signal availability in the
	buffer queue, whereas the white cells represent signal loss. The signals are read
	in a FIFO approach and fused after corresponding associations are formed based on timestamps, which
	is shown with links between the messages in the middle part of \Figure
	\ref{fig:lossy-signal-handling}. The right hand side illustrates the components of the fused signal. For example at $t_2$, the fused signals comprises of $F_L$ and $R_R$ signal only.
	
	\subsection{Lidar fusion}
	After using the prior extrinsic calibration and signal correspondences we fuse
	the lidar data from the individual sensors in the $\Base$ frame using a
	\textit{ikd-tree} \cite{xu2022fast} and downsample them to a 5 \si{cm} voxel resolution. To compensate for the motion distortion during a lidar scan, all
	points are projected to a pose corresponding to the beginning of the corresponding scan using IMU pre-integration \cite{Forster2017} between the start and end of
	the lidar scan. We use the dual quaternion (DQ) operator to handle
	interpolation of translation and rotation components together using  screw
	linear interpolation (ScLERP) \cite{kavan2005spherical}. Let, $\mathbf{p}_t$ be the the timestamp of a
	received lidar point at time $t \in [t_i, t_{i+1}]$. The point can be transformed to the beginning of the scan as $\T_{\Lidar_i\Lidar_{t}}^{-1}\mathbf{p}_t$, where, $\T_{\Lidar_i\Lidar_{t}}$ can be recovered in DQ form
	as ($\mathbf{Q}$ is the DQ form of $\T$), 
	\beq
	\begin{aligned}
		\mathbf{Q}_{\Lidar_i\Lidar_{t}} = \mathbf{Q}_{\Lidar_i} \otimes
		(\mathbf{Q}_{\Lidar_i}^{-1} \otimes \mathbf{Q}_{\Lidar_{i+1}})^{\eta}, \quad
		\eta=\frac{t-t_{i}}{t_{i+1}-t_{i}}.
	\end{aligned}
	\eeq

	\subsection{IMU fusion}
	\label{sec:MIMU}
	To create a robust IMU signal we fuse the available embedded IMUs as a set, which we call a MIMU. A naive approach for MIMU signal fusion would be averaging. Instead we use a MLE approach for the fusion \cite{skog2016inertial}.
	\begin{figure}[!h]
		\centering
		\vspace{-3mm}
		\includegraphics[width=1\columnwidth]{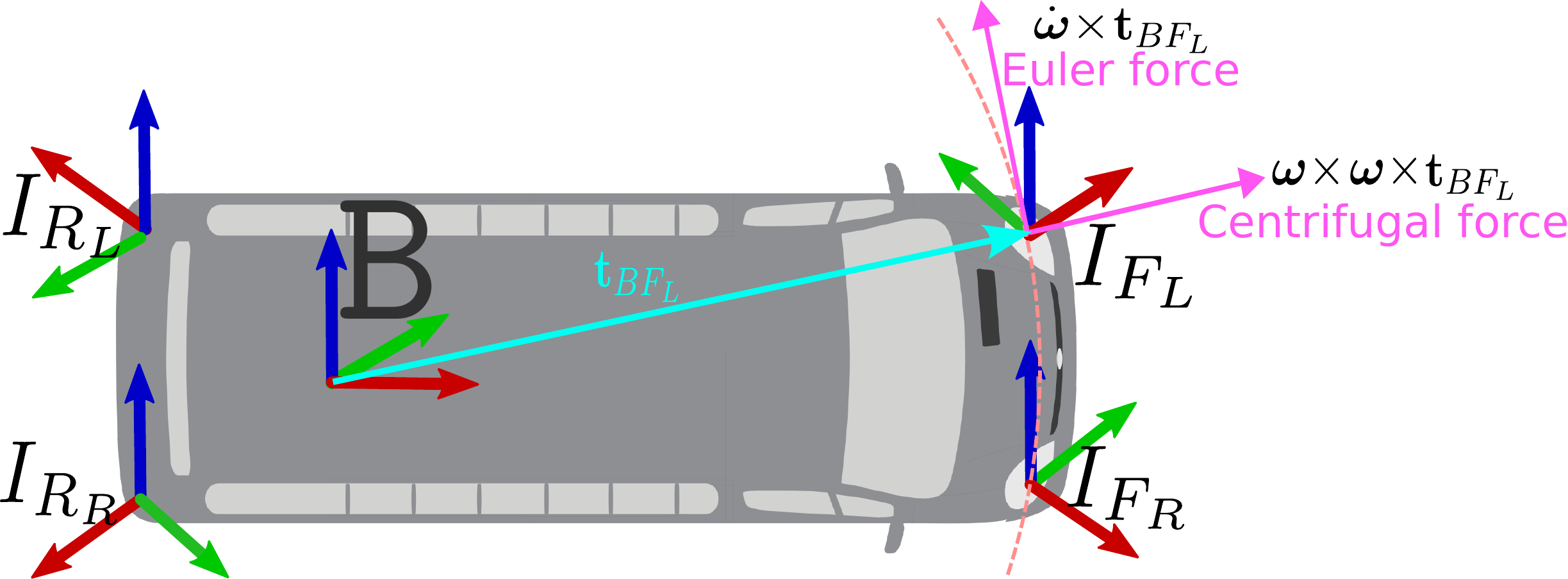}
		\vspace{-7mm}
		\caption{Illustration of IMU transformation with Coriolis force compensation.} 
		\label{fig:imu-transformation}

	\end{figure}
	\subsubsection{IMU transformation with Coriolis compensation}
	Let, $\left[\mathbf{R}_{\Base\Imu^{(k)}},\mathbf{t}_{\Base\Imu^{(k)}}\right]$ be
	the calibration parameters between the $\Base$ and the $k^{th}$ IMU frame,
	$\Imu^{(k)}$. $\mathbf{f}_{\Imu^{(k)}}, \boldsymbol{\omega}_{\Imu^{(k)}}$ are
	the linear acceleration and angular velocity measurements for the $k^{th}$ IMU
	in its corresponding sensor frame, $\Imu^{(k)}$. Before the fusion of the MIMU signals,
	all the IMU measurements must be converted to the $\Base$ frame by compensating for
	the Coriolis force illustrated in Fig. \ref{fig:imu-transformation}. Please note
	we have omitted $k$ superscript for brevity in rest of the discussion.
	\begin{equation}
		\mathbf{R}_{BI}^{-1}\mathbf{f}_{I}=\mathbf{f}_B+\underbrace{\boldsymbol{\omega}
			\times\left(\boldsymbol{\omega} \times \mathbf{t}_{BI}\right)}_{\text
			{Centrifugal force }}+\underbrace{\dot{\boldsymbol{\omega}} \times
			\mathbf{t}_{BI}}_{\text {Euler force }}
		\label{eq:coriolis}
	\end{equation} 
	\begin{equation}
		\begin{aligned}
			\mathbf{R}_{BI}^{-1}\mathbf{f}_{I}&=\mathbf{f}_B+[\boldsymbol{\omega}]_{\times}[\boldsymbol{\omega}]_{\times}
			\mathbf{t}_{BI} + [\dot{\boldsymbol{\omega}}]_{\times}\mathbf{t}_{BI}
			\\
			&= \mathbf{f}_B+[\boldsymbol{\omega}]^2_{\times} \mathbf{t}_{BI} -
			[\mathbf{t}_{BI}]_{\times}\dot{\boldsymbol{\omega}},
			\label{eq:coriolis_skew}
		\end{aligned}
	\end{equation} 
	where, $[\boldsymbol{.}]_{\times}$ is a skew-symmetric matrix and
	$[\mathbf{a}]_{\times}\mathbf{b} = - [\mathbf{b}]_{\times}\mathbf{a}$. In the
	first step of the IMU fusion, we orient all the IMU measurements in their
	respective sensor frames to $\Base$ frame by applying eq. \ref{eq:coriolis_skew}
	with suitable calibration and noise parameters, estimated in advance while characterizing the IMUs.
	\subsubsection{Accelerometer signal model} 
	From eq. \ref{eq:coriolis_skew} we can also derive the following signal model
	for an array of $K$ accelerometers, where $\mathbf{y}_f$ is the acceleration
	measurement vector and $\boldsymbol{\eta}^f$ is the measurement noise of the
	accelerometers:
	\begin{equation}
		\mathbf{y}_f=\mathbf{h}_f(\boldsymbol{\omega})+\mathbf{H}\Phi+\boldsymbol{\eta}^f,
		\label{eq:acc_model}
	\end{equation} 
	where,
	\begin{equation} 
	\resizebox{0.91\columnwidth}{!}{$
		\mathbf{h}_f(\boldsymbol{\omega})= 
		\begin{bmatrix}
			{[\boldsymbol{\omega}]^2_{\times}} \mathbf{t}_{BI^{(1)}}
			\\ 
			\vdots 
			\\ 
			[\boldsymbol{\omega}^2_{\times}] \mathbf{t}_{BI^{(K)}}
		\end{bmatrix}, 
		\mathbf{H} = 
		\begin{bmatrix}
			-[{\mathbf{t}}_{BI^{(1)}}]_{\times} &  \mathbf{I}_3
			\\ 
			\vdots & \vdots
			\\ 
			-[{\mathbf{t}}_{BI^{(K)}}]_{\times} &  \mathbf{I}_3
		\end{bmatrix},
		\Phi = 
		\begin{bmatrix}
			\dot{\boldsymbol{\omega}}
			\\
			\mathbf{f}_B
		\end{bmatrix}.
		\label{eq:acc_model_extended}$}
	\end{equation}
	
	\subsubsection{Gyroscope signal model}
	The angular velocity is sensed by gyroscope on a rigid body is the same no matter
	where the sensor is located, the same applies for angular acceleration. Hence we can
	derive a simple model for the measurement vector  $\mathbf{y}_\omega$, given the
	gyro measurement noise $\boldsymbol{\eta}^{\omega}$:
	\begin{equation}
		\begin{aligned}
			\mathbf{y}_{\omega}&=\mathbf{h}_{\omega}(\boldsymbol{\omega})+\boldsymbol{\eta}^{\omega},
			\label{eq:gyro_model}
		\end{aligned}
	\end{equation} 
	where, $\mathbf{h}_{\omega}(\boldsymbol{\omega}) = \mathbf{1}_K \otimes
		\boldsymbol{\omega}$.
	Note that in eq. \ref{eq:gyro_model} we do not model the bias, as it has been done in IMU pre-integration. Here $\mathbf{1_K}$ is a column vector of ones and
	$\otimes$ is the kronecker product. 
	
	\subsubsection{MIMU signal model}
	Concatenation of eq. \ref{eq:acc_model} and eq. \ref{eq:gyro_model} yields:
	\begin{equation}
		\mathbf{y} = \mathbf{h}(\boldsymbol{\omega}) + \mathbf{H \Phi} +
		\boldsymbol{\eta},
		\label{eq:array_model}
	\end{equation}
	where,
	\begin{equation}
	\resizebox{0.91\columnwidth}{!}{$
		\mathbf{y} =
		\begin{bmatrix}
			\mathbf{y}_f
			\\
			\mathbf{y}_{\omega}
		\end{bmatrix}, 
		\mathbf{h}(\boldsymbol{\omega}) =
		\begin{bmatrix}
			\mathbf{h}_{f}(\boldsymbol{\omega}) 
			\\
			\mathbf{h}_{\omega}(\boldsymbol{\omega}) 
		\end{bmatrix},
		\mathbf{H} = 
		\begin{bmatrix}
			\mathbf{H}_f
			\\
			\mathbf{0}_{3K,6}
		\end{bmatrix}, 
		\boldsymbol{\eta} = 
		\begin{bmatrix}
			\boldsymbol{\eta}^f
			\\
			\boldsymbol{\eta}^{\omega}
		\end{bmatrix},
		\label{eq:array_model_extended}
	$}
	\end{equation}
	
	\noindent and $\mathbf{0}_{m,n}$ denotes a zero matrix of size $m$ by $n$. 
	
	\subsubsection{Maximum likelihood estimator}
	The objective is to use eq. \ref{eq:array_model} to form a log-likelihood
	function with the assumption of zero-mean Gaussian distribution of measurement
	noises:
	\begin{equation} 
		L(\boldsymbol{\omega},\Phi)=\frac{1}{2}\left \| \mathbf{y} -
		\mathbf{h}(\boldsymbol{\omega}) - \mathbf{H \Phi} \right \|_{\mathbf{Q}^{-1}}^2.
		\label{eq:log_likelihood}
	\end{equation}
	Here, $\mathbf{Q}$ is the covariance matrix of the system and the norm is given
	by $\left \| \mathbf{X} \right \|_{\Sigma}^2 = \mathbf{X}^T\Sigma \mathbf{X}$.
	The maximum likelihood estimator is then defined as:
	\begin{align} 
		{\begin{Bmatrix}
				\boldsymbol{\omega}^{\star}, \Phi^{\star}
		\end{Bmatrix}}
		&=\underset{\boldsymbol{\omega}, \Phi}{\arg \max
		}\left[L\left(\boldsymbol{\omega}, \boldsymbol{\Phi}\right)\right].
		\label{eq:max_likelihood}
	\end{align}
	To solve eq. \ref{eq:max_likelihood}, a trivial way is to fix the
	parameter for $\boldsymbol{\omega}$, and to maximize the likelihood function with
	respect to $\Phi$, and to then replace the solution in the function. As the
	angular velocity over the sensor array is the same and independent of the
	location, we can easily calculate the fused $\boldsymbol{\omega}^{\star}$ as
	weighted least square of gyro measurements as:
	\begin{align} 
		\resizebox{0.88\columnwidth}{!}{$\boldsymbol{\omega}^{\star}=\left(\left(\mathbf{1}_{K}^{T}
			\otimes \mathbf{I}_{3}\right) \mathbf{Q}^{-1}\left(\mathbf{1}_{K} \otimes
			\mathbf{I}_{3}\right)\right)^{-1}\left(\mathbf{1}_{K}^{\top} \otimes
			\mathbf{I}_{3}\right) \mathbf{Q}^{-1} \mathbf{y}_{\boldsymbol{\omega}}$}.
		\label{eq:weighted_LS}
	\end{align}
	
	\noindent Using the estimated $\boldsymbol{\omega}^{\star}$, we can now solve
	eq. \ref{eq:max_likelihood} as follows:
	\begin{align}
		\widehat{\Phi}\left(\boldsymbol{\omega}^{\star}\right) &=\underset{\Phi}{\arg
			\max }\left[L\left(\boldsymbol{\omega}^{\star}, \Phi \right)\right]
		\nonumber\\ 
		&\stackrel{eq. \ref{eq:weighted_LS}} {=}\left(\mathbf{H}^{\top}
		\mathbf{Q}^{-1} \mathbf{H}\right)^{-1} \mathbf{H}^{\top}
		\mathbf{Q}^{-1}\left(\mathbf{y}-\mathbf{h}(\boldsymbol{\omega}^{\star})\right).
		\label{eq:imu-fusion}
	\end{align}
	Note that the right hand term represents the error, i.e. measurements vector
	$\mathbf{y}$ subtracted from the measurement model
	$\mathbf{h}(\boldsymbol{\omega}^{\star})$.
	
	\subsection{Factor graph formulation}
	A factor graph is a type of probabilistic graphical model for Bayesian
	inference, which can exploit the conditional independence between variables in
	the model and has been used widely for estimation problems in robotics
	\cite{Dellaert2017}. In our setup we optimize the intra-modal fused measurements
	in between keyframe at $t_{i}$ and $t_j$ for state estimation. Let,
	$\calI_{ij}$, $\calL_{ij}$ and $\calG_{ij}$ be the IMU, lidar and GNSS
	measurements between $t_{i}$ and $t_j$ and $\mathcal{K}$ be the set of all
	keyframes. If the measurements are assumed to be conditionally independent then
	the current state at keyframe $kf$, can be estimated as,
	\begin{multline}
		\State_{kf}^{\star} = \argmin_{\State_{kf}} \|\mathbf{r}_0\|^2_{\Sigma_0} +
		\sum_{(i,j) \in \mathcal{K}_{kf}}^{} \left(
		\|\mathbf{r}_{\calI_{ij}}\|^2_{\Sigma_{\calI_{ij}}} +
		\right. \\
		\left.
		\|\mathbf{r}_{\calL_{ij}}\|^2_{\Sigma_{\calL_{ij}}} 
		+  \|\mathbf{r}_{\calG_{i}}\|^2_{\Sigma_{\calG_{i}}}\right),
		\label{eq:factor_graph_ls}
	\end{multline}
	where, each term is the squared residual error associated to a
	factor type, weighted by the inverse of its covariance matrix. The notations:
	$\mathbf{r}_0$, $\mathbf{r}_{\calI_{ij}}$, $\mathbf{r}_{\calL_{ij}}$,
	$\mathbf{r}_{\calG_{i}}$ represent residuals for state prior, pre-integrated
	IMU, lidar odometry and GNSS factors respectively, described in the following
	sections. 
	
	\begin{figure}[!tbh]
		\centering
		\includegraphics[width=1\columnwidth]{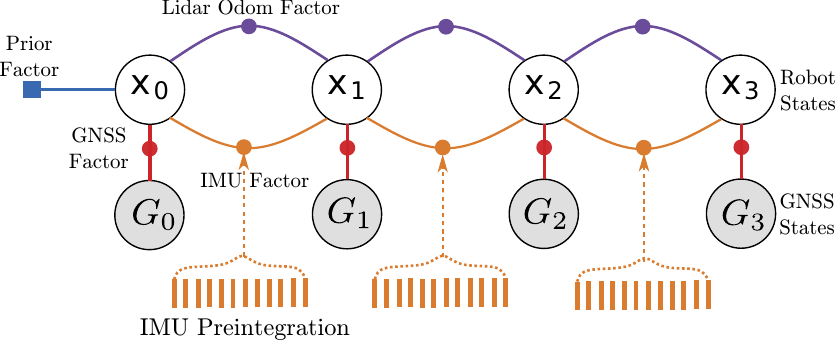}
		\vspace{-5.5mm}
		\caption{Factor graph structure with preintegrated IMU and lidar factors. GNSS poses are also added as unary factors in the graph.}
		\label{fig:factor-graph}
		\vspace{-2mm}
	\end{figure}
	
	\subsubsection{Prior factor}
	\label{subsubsec:prior-factor}
	Prior factors are used to constrain the system with the $\world$ frame. Assuming known
	prior $\tensor[_\World]{\T}{_{\mathtt{{GNSS}}_{0}}}$, we can anchor our system
	to the $\World$ frame by computing the residual between the estimated state
	$\widehat{\State}_{0}$, and GNSS based state estimate  $\tensor[_\World]{\T}{_{\mathtt{{GNSS}}_{0}}}$, at $t_0$.
	The residual becomes:
	\begin{equation}
		\mathbf{r}_0  =
		\left(\begin{array}{c}\check{\Pi}\left(\widehat{\mathbf{T}}_{0}^{-1}
			\mathbf{T}_{\mathtt{{GNSS}}_{0}}\right) \\ \widehat{\mathbf{v}}_{0}-[0\;0\;0]^{\mathtt{T}} \\
			\widehat{\mathbf{\rotvel}}_{0}- [0\;0\;0]^{\mathtt{T}} \\
			\widehat{\mathbf{b}}_{0}^{a}-\mathbf{b}_{{0}}^{a} \\
			\widehat{\mathbf{b}}_{0}^{\omega}-\mathbf{b}_{{0}}^{\omega}\end{array}\right),
	\end{equation}
	where, $\check{\Pi}$ is the lifting operator used to solve the optimization in
	tangent space according to the  ``lift-solve-retract'' principle  presented in \cite{Forster2017}.
	
	\subsubsection{IMU pre-integration factor}
	\label{subsubsec:imu-factors}
	We use the standard formulation for IMU measurement
	pre-integration \cite{Forster2017} to create the IMU factor
	between
	consecutive nodes of the pose graph. The high frequency pose
	update (at about 100Hz) is also used by the lidar odometry module. The residual has the form (with a
	derivation taken from \cite{Forster2017}):
	\begin{equation}
		\mathbf{r}_{\calI_{ij}}  = \left[ \mathbf{r}^\transpose_{\Delta
			\R_{ij}}, \mathbf{r}^\transpose_{\Delta \tran_{ij}},
		\mathbf{r}^\transpose_{\Delta \vel_{ij}},
		\mathbf{r}_{\bias^a_{ij}},
		\mathbf{r}_{\bias^g_{ij}} \right].
	\end{equation}
	
	\subsubsection{Lidar odometry and between factor}
	\label{subsubsec:lidar-odom-factors}
	\paragraph{Lidar odometry} For the lidar odometry we register the fused
	downsampled point cloud directly to local submap using \textit{ikd-tree}
	\cite{xu2022fast} leveraging its computational efficiency when finding
	correspondences. ICP is performed between corresponding points to infer
	relative motion, using a motion prior from IMU
	pre-integration. We register lidar poses at ICP frequency ($\approx$ 2 Hz),
	so as to used to maintain a smooth local submap. Since we directly register the
	raw point clouds to the local submap without feature engineering it improves the
	accuracy of the odometry system in feature-less environments. Also, as we are not specifically doing SLAM, we update the local submap periodically. This helps in managing a small memory footprint unlike maintaining a global map (which would need large memory) required for the lidar odometry. 

	
	\paragraph{Between factor formulation} Based on our optimization window we may
	sample multiple lidar odometry measurements between $t_{i}$ and $t_j$. However,
	to constrain the pose graph with a single \textit{between factor} within the smoothing
	window we use the first and the last estimated poses from
	the lidar odometry module, denoted as,
	$\widehat{\T}_{i+\epsilon}$ and $\widehat{\T}_{j-\epsilon}$ respectively. Note that we add the term $\epsilon$ to denote the
	lidar odometry module processing time. Considering $\epsilon \to 0$, the
	residual has the form:
	\begin{equation}
		\mathbf{r}_{\calL_{ij}}  = \check{\Pi}  \left(
		({\T}_{i}^{-1}{\T}_{j})^{-1}\widehat{\T}_{i}^{-1}\widehat{\T}_{j} \right).
	\end{equation}
	
	\subsubsection{GNSS prior factor}
	\label{subsubsec:gnss-prior-factors}
	Lidar inertial odometry will drift over time. Hence,
	to constrain the state estimation over long distances, we
	add only the $\tensor[_\World]{\tran}{_{\mathtt{{GNSS}}}}$ represented in UTM
	co-ordinates as a prior on our factor graph formulation if the estimated
	position covariance is larger than the GNSS position covariance adapted from
	\cite{tixiao2020lio-sam}. Note that our graph formulation will naturally produce
	state estimates even when no GNSS information is available. Since, we only use
	the translation components, the residual becomes:
	\begin{equation}
		\mathbf{r}_{\calG_{i}}  =  | \tensor[_\World]{\tran}{_{\mathtt{{GNSS}}_i}} -
		\tensor[_\World]{\tran}{_{\mathtt{{\Base}}_i}}|.
	\end{equation}
	
	For simplicity, we have used constant values for the covariances of IMU
	pre-integration factor, lidar \textit{between factor} and GNSS prior factor for our
	experiments.

	\section{Experimental Results}
	To demonstrate system performance, we conducted several experiments in real world environments including urban and highway driving scenarios to collect data using two vehicles with the sensor setup illustrated in \Figure\ref{fig:coordinate-frames}.

	
	\subsection{Dataset}
	\label{sec:dataset}
	All sensing data in our vehicle is synchronized using PTP (Precision Time Protocol).
	For ground truth (GT) pose estimates, we use data from a GNSS receiver. 

	We use the following notation to distinguish between different sensor configurations -- $\mathbf{L_{n_l}I_{n_i}G_{n_g}}$ (L: Lidar, I: IMU and G: GNSS); with $\mathbf{n_l}$, $\mathbf{n_i}$ and $\mathbf{n_g}$: indicating the corresponding number of sensor. 

	We collected 5 test sequences which is described in details in Table~\ref{tab:dataset}. Lossy signal conditions were emulated manually for our experiments by dropping individual sensor signals randomly once for a period of 5-10 seconds in two data logs. For the single modality experiments, we use the $F_L$ lidar and its corresponding IMU in both the bus and the truck logs.
	\begin{table}[!h]
		\centering
		\vspace{2mm}
		\fontsize{18}{18}\selectfont
		\resizebox{\columnwidth}{!}{
			\begin{tabular}{l|llcccc}  \toprule
				\multicolumn{7}{c}{Datasets collection details for the experiments}\\
				\midrule \midrule
				\textbf{Data} & Vehicle & Scenario & Sensor setup & Length (\si{Km}) & Duration (\si{secs}) & Signal loss\\
				\midrule
				Seq-1 & Bus & Urban & ${L_4I_4G_1}$ & 1.643 & 333.37 & Yes\\
				Seq-2\textsuperscript{\dag} & Bus & Highway & ${L_4I_4G_1}$ &  1.948 & 168.64 & No\\
				Seq-3\textsuperscript{\dag} & Bus & Highway & ${L_4I_4G_1}$ & 1.577 & 118.56 & No\\
				Seq-4 & Truck & Urban & ${L_4I_4G_1}$ & 0.453 & 78.91 & No\\
				Seq-5 & Truck & Urban & ${L_4I_4G_1}$ & 1.267 & 151.37 & Yes\\
				\bottomrule
				\multicolumn{7}{l}{\textsuperscript{\dag}\textit{Four front lidars and IMUs are used for this log sequence.}}
		\end{tabular}}
	\vspace{-3mm}
		\caption{}
		\label{tab:dataset}
		\vspace{-4mm}
	\end{table}
We demonstrate the lidar fusion method and its results in signal dropout scenario in the supplementary video.

	\subsection{IMU fusion}
\begin{figure}[!t]
		\centering
		\vspace{2mm}
		\includegraphics[width=1\columnwidth]{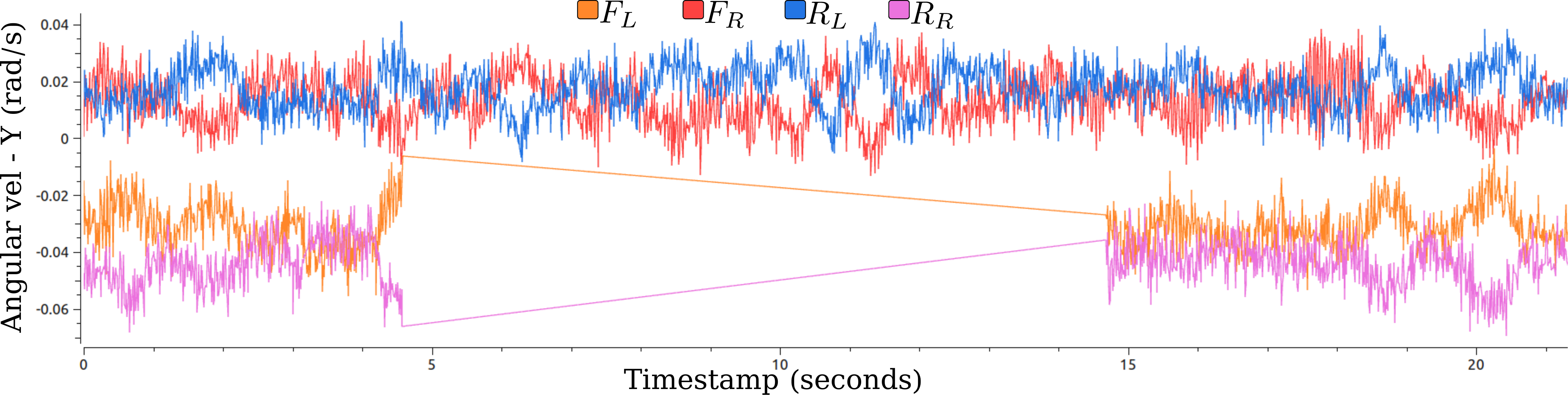}
		\vspace{-6mm}
		\caption{IMU raw signal of angular velocity (Y-component).}
		\label{fig:imu-raw}
		\vspace{1mm}
	\end{figure}

	All the IMUs from the integrated sensor housing (shown in \Figure\ref{fig:coordinate-frames}) are oriented along the $\Base$ frame using the predefined calibration parameters. Due to limited space availability, we only visualize and discuss the results of the fusion of the 
	angular velocity (Y-component) for Seq-1 and confirm that a similar hypothesis holds true for the other IMU signal components in all the sequences. 
	As seen in Fig. \ref{fig:imu-raw}, the angular velocity components look out-of-phase, since they are not transformed to the $\Base$ frame. Also, around fifth second in Fig. \ref{fig:imu-raw}, we observe that there was a signal loss for the $F_L$ and $R_R$ IMU for around 10 seconds. 		

\begin{figure}[!t]
		\centering
		\vspace{-1mm}
		\includegraphics[width=1\columnwidth]{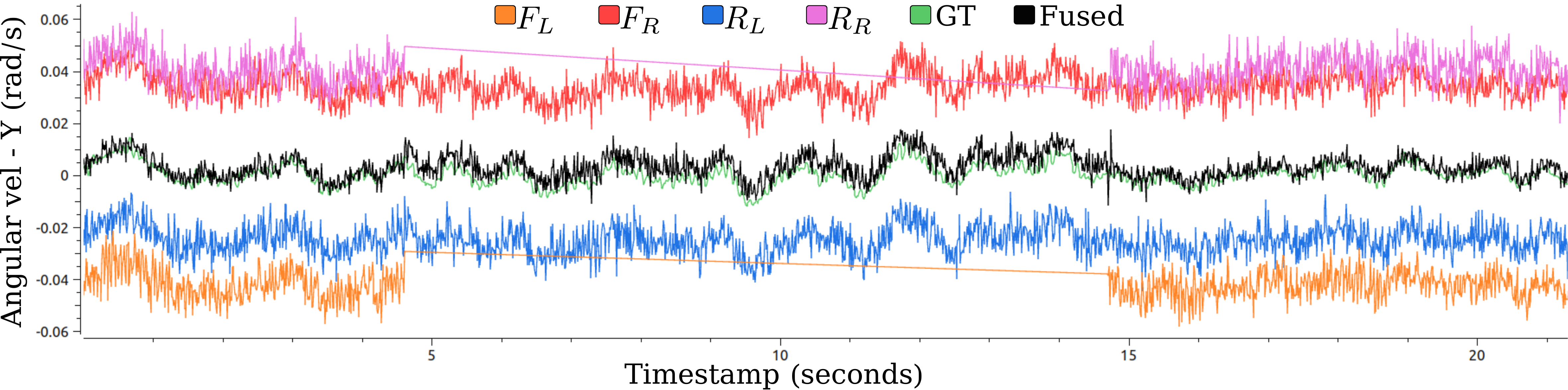}
		\vspace{-6mm}
		\caption{IMU fusion of angular velocity (Y-component).}
		\label{fig:imu-fusion}
		\vspace{-5mm}
	\end{figure}

	After applying the MIMU fusion technique, described in Sec.~\ref{sec:MIMU}, with the correct calibration parameters we get a fused MIMU signal in $\Base$ frame, which is robust to signal loss as seen in 
	\Figure \ref{fig:imu-fusion}. We also analyzed the fused signal \textit{wrt} GT and verify the quality of the fusion. The GT IMU signal is logged from the GNSS system in $\Base$ frame. As seen in 
	\Figure \ref{fig:imu-fusion} our fused signal for the 
	angular velocity (Y-component) closely corresponds to its respective GT.
	
	\begin{table}[!h]
		\vspace{-1mm}
		\centering
		\resizebox{\columnwidth}{!}{
			\begin{tabular}{l|cccc}  \toprule
				\multicolumn{5}{c}{\textbf{RMSE} of IMU Fusion \textit{wrt} Ground Truth}\\
				\midrule \midrule
				\multirow{2}{*}{\raisebox{-\heavyrulewidth}{\textbf{Data}}} &
				\multicolumn{2}{c}{Average Fusion} & \multicolumn{2}{c}{MLE Fusion}
				\\
				\cmidrule(lr){2-5}
				& Linear acceleration & Angular velocity & Linear acceleration & Angular velocity \\
				\midrule
				Seq-1 & 2.868 & 0.004 & \textbf{1.716} & \textbf{0.003}\\
				Seq-2 & 2.219 & 0.003 & \textbf{1.814} & \textbf{0.003}\\
				Seq-3 & 2.186 & 0.003 & \textbf{1.792} & \textbf{0.003}\\
				Seq-4 & 2.041 & 0.008 & \textbf{1.818} & \textbf{0.007}\\
				Seq-5 & 2.263 & 0.008 & \textbf{1.820} & \textbf{0.007}\\
				\bottomrule
		\end{tabular}}
		\caption{}
		\label{tab:imu_ls_vs_avg}
		\vspace{-8mm}
	\end{table}
	To provide more detailed analysis we summarize our RMSE error for the fused signal \textit{wrt} the GT signal and compare its performance to a fusion approach based on averaging of the signals in Tab \ref{tab:imu_ls_vs_avg}. For the RMSE, we compared the linear acceleration and the angular velocity components separately as, $\operatorname{RMSE} = \sqrt{\frac{1}{N} \|I_{GT} - \widehat{I}\|^2_F}$.

	\begin{figure*}[!h]
	\centering
	\vspace{2mm}
	\begin{multicols}{2}
		\centering
		\includegraphics[width=1\columnwidth]{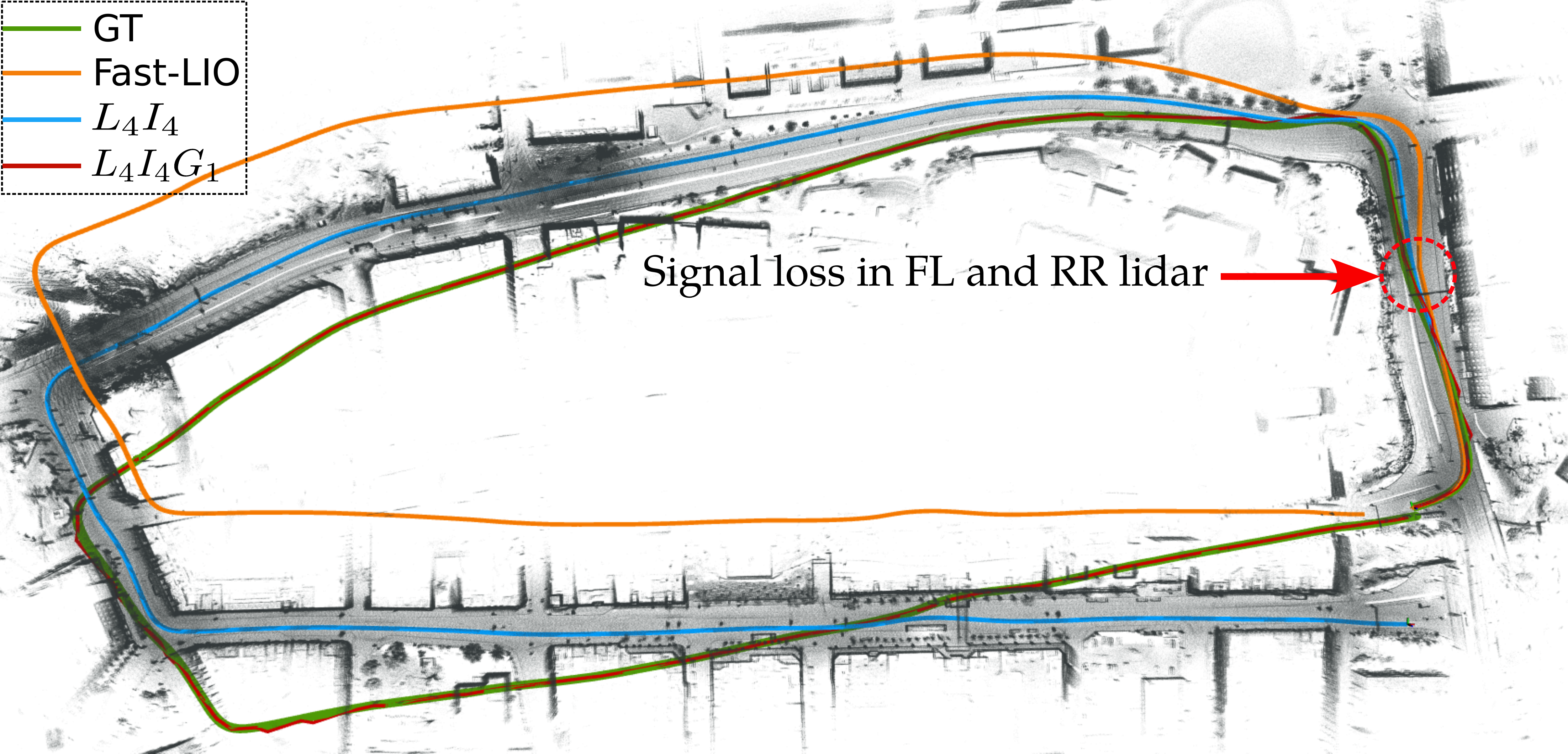}
		\vspace{-10mm}
		\caption{Seq 1: Driving scenario with $F_L$ and $R_R$ lidar signal loss (Bus)}
		\label{fig:seq_1_gnss_5_hz}
		\vspace{6.5mm}
		\includegraphics[width=1\columnwidth]{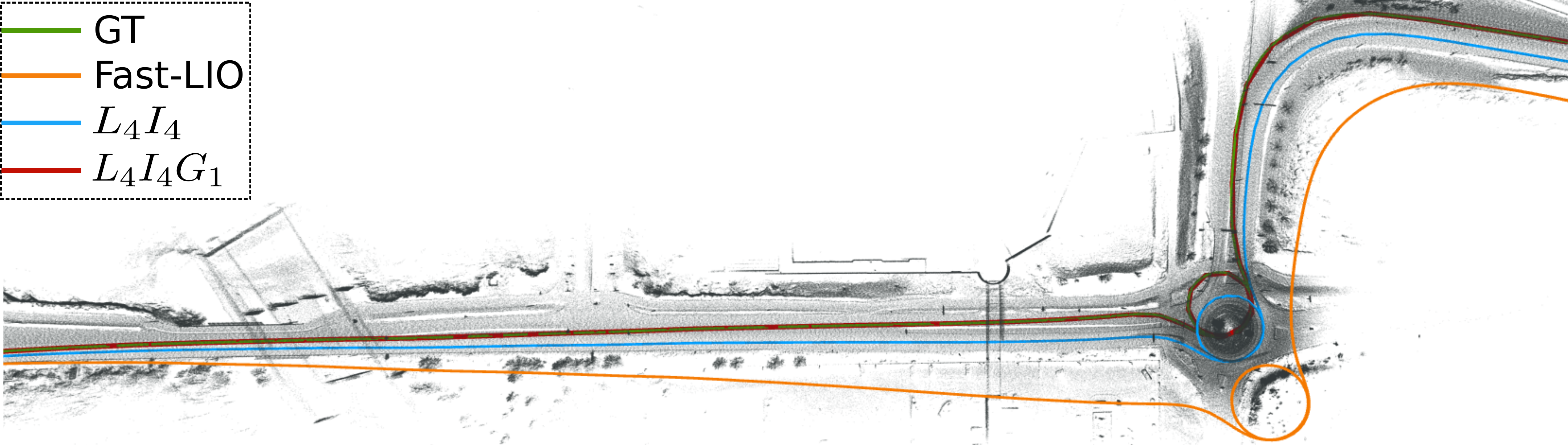}
		\caption{Seq 2: Driving scenario with a roundabout (Bus)}
		\label{fig:seq_2_gnss_5_hz}
		\vspace{-4mm}
		\includegraphics[width=1\columnwidth]{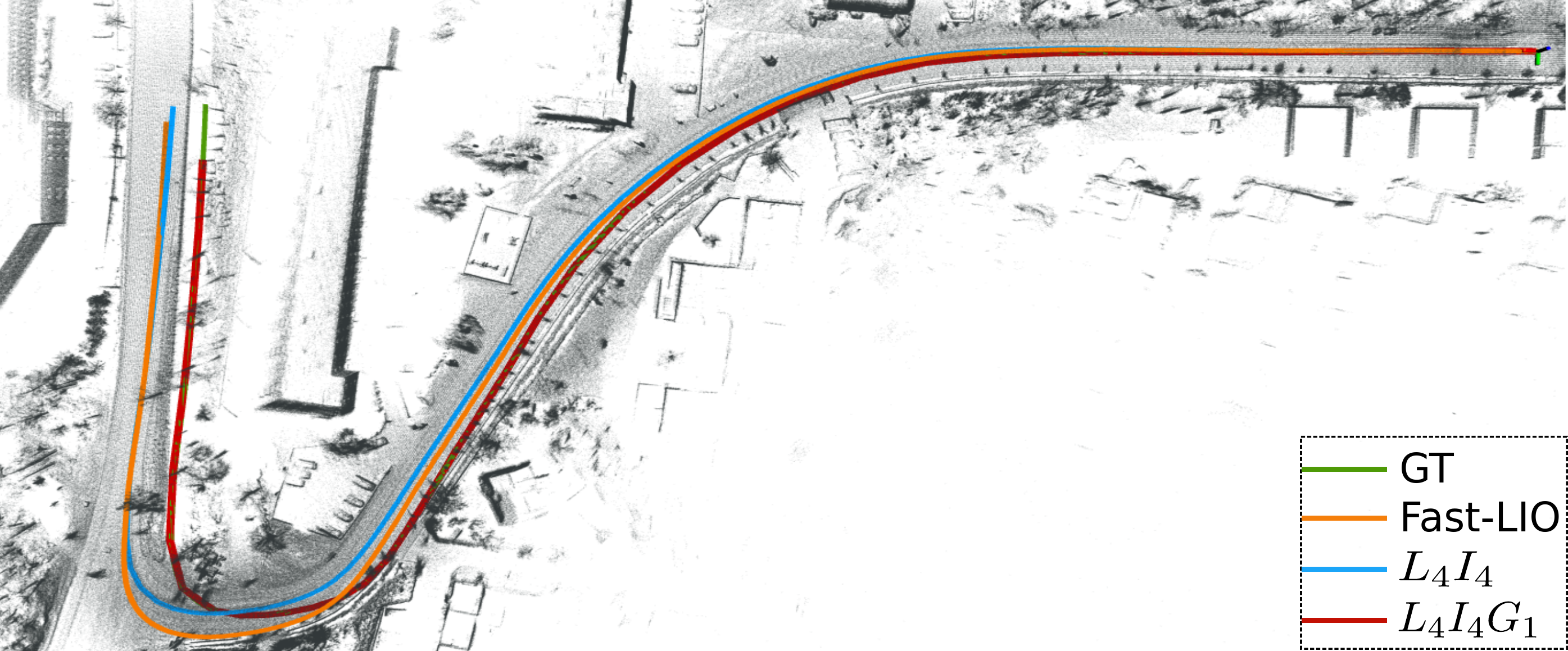}
		\caption{Seq 4: Driving scenario with long turning radius (Truck)}
		\label{fig:seq_4_gnss_5_hz}
		\includegraphics[width=1\columnwidth]{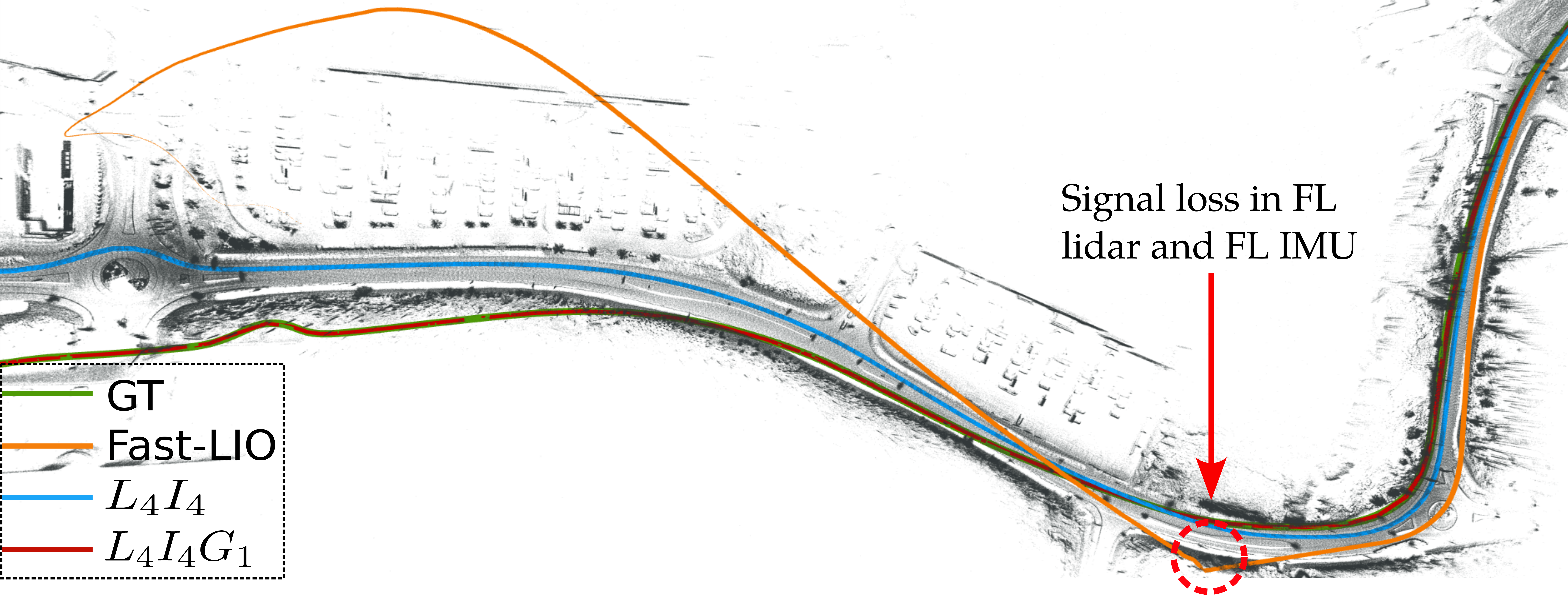}
		\vspace{-6mm}
		\caption{Seq 5: Driving scenario with $F_L$ lidar and IMU signal loss (Truck)}
		\label{fig:seq_5_gnss_5_hz}
	\end{multicols}
\vspace{-2mm}
\end{figure*}

	\subsection{State Estimation}	
%

	The relative pose error (RPE) and absolute pose error (APE) metrics are used to evaluate the local and global consistency respectively of the estimated poses. RPE compares the relative pose along the estimated and the reference
	trajectory between two timestamps $t_i$ and $t_j$, using the relative transform over a specific distance traveled (10 \si{m}) as, 
	\begin{align}
		\operatorname{RPE}_{i,j} =({\T}_{GT_{i}}^{-1}{\T}_{GT_{j}})^{-1}(\widehat{{\T}}{_{i}^{-1}}\widehat{{\T}}{_{j}}).
	\end{align}
	APE is based on the absolute relative pose difference between two poses
	${\T}{_{GT_i}}$ and $\widehat{{\T}}{_i}$ at timestamp $t_i$ and is compared for the whole trajectory of $N$ timestamps as, 
	\begin{align}
		\operatorname{APE}_N = \sqrt{\frac{1}{N}\sum_{i=1}^{N}\|{\T}{_{GT_i}^{-1}}\widehat{\T}{_i} - I_{4\times4}\|^2_F}.
	\end{align}

	In our experiments to demonstrate the performance improvement of using multiple
	sensors versus a single sensor when signal loss occurs 
	we used the $L_4I_4$ configuration, described in Sec. \ref{sec:dataset}. To illustrate the reduction in drift when additionally fusing GNSS, we used the $L_4I_4G_1$ configuration.
	
	As seen in \Figure\ref{fig:imu-raw}, signal loss for few seconds occurs in Seq-1 (\Figure\ref{fig:seq_1_gnss_5_hz}). Fast-LIO2 drifts quickly at that point --- but maintains consistency in the rest of the trajectory. Meanwhile, the $L_4I_4$ configuration handles this situation seamlessly using the other available sensor signals. 

	In the highway scenario of Seq-2 (\Figure\ref{fig:seq_2_gnss_5_hz}), we 
	see that again $L_4I_4$ configuration outperforms Fast-LIO2 as there are very few environmental structures. Hence, expanding the lidar field-of-view using $L_4I_4$ gives more constraints for the ICP matching, outperforming Fast-LIO2.
	
	In Seq-5 (\Figure\ref{fig:seq_5_gnss_5_hz}), we manually induced a data drop in the $F_L$ lidar and IMU for 8 seconds. Fast-LIO2 again fails to recover from the situation, whereas, $L_4I_4$ configuration performs well again.

	\begin{table}[!h]
		\centering
		\vspace{3mm}
		\resizebox{\columnwidth}{!}{
			\begin{tabular}{l|ccc}  \toprule
				\multicolumn{4}{c}{\textbf{Pose Error} -- \textbf{RPE} 
					\textbf{(}Translation[\si{\metre}], Rotation [\si{\degree}]\textbf{)}
					/ \textbf{APE}} \\
				\midrule \midrule
				\textbf{Data} & $\mathbf{L_4I_4}$& $\mathbf{L_4I_4G_1}$ &
				$\mathbf{Fast-LIO2}\textsuperscript{\dag}$\\
				\midrule
				Seq-1 & 0.676, \textbf{1.248} / 4.992 & \textbf{0.115}, 1.719 / \textbf{3.574} & 0.977, 1.458 / 11.137\\
				Seq-2 & 0.419, \textbf{0.264} / 5.376 & \textbf{0.158}, 0.804 / \textbf{4.342} & 0.673, 0.458 / 24.422\\
				Seq-3 & 0.361, \textbf{0.228} / 3.479 & \textbf{0.221}, 0.655 / \textbf{2.268} & 0.421, 0.298 / 13.770\\
				Seq-4 & 0.835, 0.471 / 1.446 & \textbf{0.106}, 0.719 / \textbf{0.785} & 0.824, \textbf{0.457} / 0.9840\\
				Seq-5 & 1.265, \textbf{0.454} / 9.556 & \textbf{0.605}, 1.479 / \textbf{1.645} & 26.815, 2.238 / 4613.7\\
				\bottomrule
				\multicolumn{4}{l}{\textsuperscript{\dag}\textit{Fast-LIO2 $ = {L_1I_1}$ configuration.}} \\
		\end{tabular}}
		\vspace{-4mm}
		\caption{}
		\vspace{-8mm}
		\label{tab:pe}
	\end{table}
	Table \ref{tab:pe}, show RPE and APE metrics for each sequence with Fast-LIO2 acting as the baseline. 
	In all the scenarios, the $L_4I_4G_1$ configuration outperform the other methods, as GNSS pose priors are added seamlessly in a graph based optimization, which constrains the pose drift but suffers from discontinuities; thus for visualization of the lidar map we use the $L_4I_4$ lidar odometry frame. 	In most of the scenarios the RPE and APE for $L_4I_4$ configuration is less than Fast-LIO2, since it encompasses a wider field-of-view with more matching constraints for ICP. However, in Seq-4 (\Figure\ref{fig:seq_4_gnss_5_hz}) Fast-LIO2 is performing slightly better as the vehicle is driven in a structure rich environment with no signal dropouts. For Seq-3 results please check our supplementary video.

	\section{Conclusion}
	\label{sec:conclusion}
	We have presented a method for multi-lidar and MIMU state estimation in a factor graph optimization framework and additionally demonstrated the use of GNSS priors. The proposed method produces comparable performance to a state-of-the-art lidar-inertial odometry system (Fast-LIO2) but outperforms it when there are signal dropouts with an average improvement of 61\% in relative translation and 42\% in relative rotational error. For future work we aim to introduce radar as an additional modality to discard dynamic environmental points for improving the odometry estimate.

	\section{Acknowledgements}
	This research has been jointly funded by the Swedish Foundation for Strategic Research (SSF) and Scania. The authors would like to thank Ludvig af Klinteberg for his valuable comments which improved the manuscript and Goksan Isil for help with experimental data creation.
	
	
	
	\bibliographystyle{./IEEEtran}
	\bibliography{./IEEEabrv, ./library}
	
	
\end{document}

%% file: shortcuts.tex
\newcommand{\hide}[1]{}
\newcommand{\Figure}{Fig.~}

\newcommand{\bdmath}{\begin{dmath}}
\newcommand{\edmath}{\end{dmath}}
\newcommand{\beq}{\begin{equation}}
\newcommand{\eeq}{\end{equation}}
\newcommand{\bdm}{\begin{displaymath}}
\newcommand{\edm}{\end{displaymath}}
\newcommand{\bea}{\begin{eqnarray}}
\newcommand{\eea}{\end{eqnarray}}
\newcommand{\beal}{\beq \begin{array}{ll}}
\newcommand{\eeal}{\end{array} \eeq}
\newcommand{\beas}{\begin{eqnarray*}}
\newcommand{\eeas}{\end{eqnarray*}}
\newcommand{\ba}{\begin{array}}
\newcommand{\ea}{\end{array}}
\newcommand{\bit}{\begin{itemize}}
\newcommand{\eit}{\end{itemize}}
\newcommand{\ben}{\begin{enumerate}}
\newcommand{\een}{\end{enumerate}}

\newcommand{\Real}{\mathbb{R}}

\newcommand{\SEthree}{\ensuremath{\mathrm{SE}(3)}\xspace}

\newcommand{\calG}{{\cal G}}

\newcommand{\calI}{{\cal I}}

\newcommand{\calL}{{\cal L}}

\newcommand{\T}{\mathbf{T}}
\newcommand{\R}{\mathbf{R}}

\newcommand{\transpose}{\mathsf{T}}

\newcommand{\rotvel}{\boldsymbol\omega}

\newcommand{\tran}{\mathbf{t}}

\newcommand{\vel}{\mathbf{v}}

\newcommand{\bias}{\mathbf{b}}

\newcommand{\State}{\boldsymbol{x}}

\newcommand{\World}{\mathtt{W}}
\newcommand{\Imu}{\mathtt{I}}
\newcommand{\Camera}{\mathtt{C}}
\newcommand{\Lidar}{\mathtt{L}}
\newcommand{\GNSS}{\mathtt{G}}
\newcommand{\world}{\mathtt{{W}}}

\newcommand{\imu}{\mathtt{{I}}}
\newcommand{\base}{\mathtt{{B}}}

\newcommand{\Base}{\mathtt{{B}}}

\SetCommentSty{algo}
\SetKwInput{KwInput}{Input} 
\SetKwInput{KwOutput}{Output} 

\newcommand{\etalcite}[2]{#1~et~al.~\cite{#2}}

\DeclareMathOperator*{\argmin}{arg\,min}

\newcommand\cancel{\bgroup\markoverwith{\textcolor{red}{\rule[0.5ex]{2pt}{0.4pt}}}\ULon}


%% file: paper.bbl
\begin{thebibliography}{10}
\providecommand{\url}[1]{#1}
\csname url@rmstyle\endcsname
\providecommand{\newblock}{\relax}
\providecommand{\bibinfo}[2]{#2}
\providecommand\BIBentrySTDinterwordspacing{\spaceskip=0pt\relax}
\providecommand\BIBentryALTinterwordstretchfactor{4}
\providecommand\BIBentryALTinterwordspacing{\spaceskip=\fontdimen2\font plus
\BIBentryALTinterwordstretchfactor\fontdimen3\font minus
  \fontdimen4\font\relax}
\providecommand\BIBforeignlanguage[2]{{%
\expandafter\ifx\csname l@#1\endcsname\relax
\typeout{** WARNING: IEEEtran.bst: No hyphenation pattern has been}%
\typeout{** loaded for the language `#1'. Using the pattern for}%
\typeout{** the default language instead.}%
\else
\language=\csname l@#1\endcsname
\fi
#2}}

\bibitem{lic-fusion}
X.~{Zuo}, P.~{Geneva}, W.~{Lee}, Y.~{Liu}, and G.~{Huang}, ``{LIC-Fusion}:
  {LiDAR}-inertial-camera odometry,'' in \emph{IEEE/RSJ International
  Conference on Intelligent Robots and Systems}, 2019, pp. 5848--5854.

\bibitem{shan2021lvi}
T.~Shan, B.~Englot, C.~Ratti, and D.~Rus, ``{LVI-SAM}: Tightly-coupled
  lidar-visual-inertial odometry via smoothing and mapping,'' in \emph{IEEE
  International Conference on Robotics and Automation}, 2021, pp. 5692--5698.

\bibitem{wisth2020vilens}
D.~Wisth, M.~Camurri, and M.~Fallon, ``{VILENS}: Visual, inertial, lidar, and
  leg odometry for all-terrain legged robots,'' \emph{IEEE Transactions on
  Robotics}, pp. 1--18, 2022.

\bibitem{sun2019scalability}
P.~Sun, H.~Kretzschmar, X.~Dotiwalla, A.~Chouard, V.~Patnaik, P.~Tsui, J.~Guo,
  Y.~Zhou, Y.~Chai, B.~Caine, \emph{et~al.}, ``Scalability in perception for
  autonomous driving: Waymo open dataset,'' in \emph{IEEE Conference on
  Computer Vision and Pattern Recognition}, 2020, pp. 2446--2454.

\bibitem{geyer2020a2d2}
J.~Geyer, Y.~Kassahun, M.~Mahmudi, X.~Ricou, R.~Durgesh, A.~S. Chung,
  L.~Hauswald, V.~H. Pham, M.~M{\"u}hlegg, S.~Dorn, T.~Fernandez,
  M.~J{\"a}nicke, S.~Mirashi, C.~Savani, M.~Sturm, O.~Vorobiov, M.~Oelker,
  S.~Garreis, and P.~Schuberth, ``{A2D2: Audi Autonomous Driving Dataset},''
  \url{https://www.a2d2.audi}, 2020.

\bibitem{palieri2020locus}
M.~Palieri, B.~Morrell, A.~Thakur, K.~Ebadi, J.~Nash, A.~Chatterjee,
  C.~Kanellakis, L.~Carlone, C.~Guaragnella, and A.~Agha-Mohammadi, ``Locus: A
  multi-sensor lidar-centric solution for high-precision odometry and 3d
  mapping in real-time,'' \emph{IEEE Robotics and Automation Letters}, vol.~6,
  no.~2, pp. 421--428, 2020.

\bibitem{jiao2021robust}
J.~Jiao, H.~Ye, Y.~Zhu, and M.~Liu, ``Robust odometry and mapping for
  multi-lidar systems with online extrinsic calibration,'' \emph{IEEE
  Transactions on Robotics}, vol.~38, no.~1, pp. 351--371, 2021.

\bibitem{skog2016inertial}
I.~Skog, J.~O. Nilsson, P.~H{\"a}ndel, and A.~Nehorai, ``Inertial sensor
  arrays, maximum likelihood, and cram{\'e}r--rao bound,'' \emph{IEEE
  Transactions on Signal Processing}, vol.~64, no.~16, pp. 4218--4227, 2016.

\bibitem{faizullin2021best}
M.~Faizullin and G.~Ferrer, ``Best axes composition: Multiple gyroscopes imu
  sensor fusion to reduce systematic error,'' in \emph{IEEE European Conference
  on Mobile Robots}, 2021, pp. 1--7.

\bibitem{Zhang2017}
J.~Zhang and S.~Singh, ``{Low-drift and real-time lidar odometry and
  mapping},'' \emph{Autonomous Robots}, vol.~41, no.~2, pp. 401--416, Feb.
  2017.

\bibitem{segal2009generalized}
A.~Segal, D.~Haehnel, and S.~Thrun, ``Generalized-{ICP},'' in \emph{Robotics:
  science and systems}, vol.~2, no.~4.\hskip 1em plus 0.5em minus 0.4em\relax
  Seattle, WA, 2009, p. 435.

\bibitem{shan-legoloam}
T.~{Shan} and B.~{Englot}, ``{LeGO-LOAM}: Lightweight and ground-optimized
  lidar odometry and mapping on variable terrain,'' in \emph{IEEE/RSJ
  International Conference on Intelligent Robots and Systems}, 2018, pp.
  4758--4765.

\bibitem{tixiao2020lio-sam}
T.~Shan, B.~Englot, D.~Meyers, W.~Wang, C.~Ratti, and D.~Rus, ``Lio-sam:
  Tightly-coupled lidar inertial odometry via smoothing and mapping,'' in
  \emph{IEEE/RSJ International Conference on Intelligent Robots and Systems},
  2020, pp. 5135--5142.

\bibitem{engel2014lsd}
J.~Engel, T.~Sch{\"o}ps, and D.~Cremers, ``{LSD-SLAM}: Large-scale direct
  monocular slam,'' in \emph{European Conference on computer vision}, 2014, pp.
  834--849.

\bibitem{xu2022fast}
W.~Xu, Y.~Cai, D.~He, J.~Lin, and F.~Zhang, ``{FAST-LIO2}: Fast direct
  lidar-inertial odometry,'' \emph{IEEE Transactions on Robotics}, vol.~38,
  no.~4, pp. 2053--2073, 2022.

\bibitem{bai2022faster}
C.~Bai, T.~Xiao, Y.~Chen, H.~Wang, F.~Zhang, and X.~Gao, ``{Faster-LIO}:
  Lightweight tightly coupled lidar-inertial odometry using parallel sparse
  incremental voxels,'' \emph{IEEE Robotics and Automation Letters}, vol.~7,
  no.~2, pp. 4861--4868, 2022.

\bibitem{furgale2014representing}
P.~Furgale, ``Representing robot pose: The good, the bad, and the ugly,'' in
  \emph{Workshop on Lessons Learned from Building Complex Systems, IEEE
  International Conference on Robotics and Automation}, 2014.

\bibitem{pedley2013tilt}
M.~Pedley, ``Tilt sensing using a three-axis accelerometer,'' \emph{Freescale
  semiconductor application note}, vol.~1, pp. 2012--2013, 2013.

\bibitem{allan1966statistics}
D.~W. Allan, ``Statistics of atomic frequency standards,'' \emph{Proceedings of
  the IEEE}, vol.~54, no.~2, pp. 221--230, 1966.

\bibitem{Forster2017}
C.~Forster, L.~Carlone, F.~Dellaert, and D.~Scaramuzza, ``On-manifold
  preintegration for real-time visual-inertial odometry,'' \emph{IEEE
  Transactions on Robotics}, vol.~33, no.~1, pp. 1--21, 2017.

\bibitem{kavan2005spherical}
L.~Kavan and J.~{\v{Z}}{\'a}ra, ``Spherical blend skinning: a real-time
  deformation of articulated models,'' in \emph{Symposium on Interactive 3D
  graphics and games}, 2005, pp. 9--16.

\bibitem{Dellaert2017}
F.~Dellaert and M.~Kaess, ``Factor graphs for robot perception,''
  \emph{Foundations and Trends in Robotics}, vol.~6, pp. 1--139, Aug. 2017.

\end{thebibliography}
